\documentclass[twocolumn, switch]{article} 

\usepackage{preprint}

\usepackage{siunitx}
\usepackage{amsmath}
\usepackage{amssymb}
\usepackage{mathtools}
\usepackage{subcaption}
\usepackage{pgfplots}

\usepackage[numbers,square]{natbib}
\usepackage[utf8]{inputenc}	
\usepackage[T1]{fontenc}	
\usepackage{xcolor}		
\usepackage[colorlinks = true,
            linkcolor = purple,
            urlcolor  = blue,
            citecolor = cyan,
            anchorcolor = black]{hyperref}	
\usepackage{booktabs} 		
\usepackage{nicefrac}		
\usepackage{microtype}		
\usepackage{lineno}		
\usepackage{float}			

\usepackage{lipsum}		

\usepackage{newfloat}
\DeclareFloatingEnvironment[name={Supplementary Figure}]{suppfigure}
\usepackage{sidecap}
\sidecaptionvpos{figure}{c}

\usepackage{titlesec}
\titlespacing\section{0pt}{12pt plus 3pt minus 3pt}{1pt plus 1pt minus 1pt}
\titlespacing\subsection{0pt}{10pt plus 3pt minus 3pt}{1pt plus 1pt minus 1pt}
\titlespacing\subsubsection{0pt}{8pt plus 3pt minus 3pt}{1pt plus 1pt minus 1pt}

\usepackage{tikz,xcolor,hyperref}

\title{Physics-Informed Learning for Time-Resolved Angiographic Contrast Agent Concentration Reconstruction}

\usepackage{authblk}

\author[1,2,*]{Noah Maul}
\author[2]{Annette Birkhold}
\author[1]{Fabian Wagner}
\author[1]{Mareike Thies}
\author[1]{Maximilian Rohleder}
\author[3,4]{Philipp Berg}
\author[2]{Markus Kowarschik}
\author[1]{Andreas Maier}

\affil[1]{Pattern Recognition Lab, Department of Computer Science, Friedrich-Alexander Universität Erlangen-Nürnberg, Germany}

\affil[2]{Siemens Healthineers AG, Forchheim, Germany}

\affil[3]{Research Campus STIMULATE, University of Magdeburg, Germany}
\affil[4]{Department of Medical Engineering, University of Magdeburg, Germany}
\affil[*]{\textit{noah.maul@fau.de}}

\begin{document}

\twocolumn[ 
  \begin{@twocolumnfalse} 
  
\maketitle

\begin{abstract}
Three-dimensional Digital Subtraction Angiography (3D-DSA) is a well-established X-ray-based technique for visualizing vascular anatomy. 
Recently, four-dimensional DSA (4D-DSA) reconstruction algorithms have been developed to enable the visualization of volumetric contrast flow dynamics through time-series of volumes. . 
This reconstruction problem is ill-posed mainly due to vessel overlap in the projection direction and geometric vessel foreshortening, which leads to information loss in the recorded projection images.
However, knowledge about the underlying fluid dynamics can be leveraged to constrain the solution space. 
In our work, we implicitly include this information in a neural network-based model that is trained on a dataset of  image-based blood flow simulations.
The model predicts the spatially averaged contrast agent concentration for each centerline point of the vasculature over time, lowering the overall computational demand.
The trained network enables the reconstruction of relative contrast agent concentrations with a mean absolute error of \SI{0.02(0.02)}{} and a mean absolute percentage error of \SI{5.31(9.25)}{\percent}.
Moreover, the network is robust to varying degrees of vessel overlap and vessel foreshortening.
Our approach demonstrates the potential of the integration of machine learning and blood flow simulations in time-resolved 
angiographic flow reconstruction.
\end{abstract}
\vspace{0.35cm}

  \end{@twocolumnfalse} 
] 



\section{Introduction}
\sloppy
Optimal diagnosis and treatment of vascular abnormalities require a detailed understanding of the present pathology including its hemodynamic properties. 
For complex vascular abnormalities, such as intracranial aneurysms or arteriovenous malformations, imaging modalities with high spatial and temporal resolution are required. 
Digital Subtraction Angiography (DSA) is an established X-ray-based imaging technique for visualizing vascular anatomy. In addition to the acquisition of 2D projection images from a single view (2D-DSA), modern angiography systems allow the reconstruction of subtracted volumetric images using cone-beam computed tomography (3D-DSA).
Although static 3D-DSA images can resolve the vasculature in 3D,  they do not provide information about dynamic blood flow as a series of 2D-DSA images \cite{Hoffman_2021}. However, due to the single view, 2D-DSA images are impaired by vascular overlay and vessel foreshortening, limiting their application for complex abnormalities.
This has motivated the development of flow reconstruction algorithms that estimate a time series of contrast intensities from a single rotational acquisition \cite{mistretta_2011,Davis_2013}.

The time series estimation is considered an ill-posed inverse problem, as single 2D projection images cannot capture the full 3D vascular filling at each timestep, which would be necessary for fully sampling a 4D reconstruction solely from measurements.
Davis et al. \cite{Davis_2013} introduced a  4D-DSA reconstruction technique that provides a series of time-resolved vascular volumes derived
from the projection images of a conventional 3D-DSA rotational acquisition. 
Here, contrast agent (CA) is injected at or shortly after the beginning of a rotational acquisition and projection images are recorded at a certain frame rate.
Afterwards, a prior static 3D-DSA reconstruction is performed, which acts as a constraint for the 4D-DSA algorithm and enables the reconstruction of the time-resolved volumetric filling. 
Each projection image corresponds to one time point and vascular filling state.
However, the reconstructed 4D image quality is compromised whenever vessel overlap is present in the acquired projection data. When the projection images are backprojected into a static 3D constraining volume, the measured attenuation cannot be uniquely attributed to the individual vessels.
This introduces artifacts corresponding to vessel segments that appear as contrasted either too early or too late (in time), decreasing the image quality of time-resolved 3D images. 
To minimize this issue, a 4D-DSA reconstruction algorithm employing a physically motivated and plausibility-based flow constraint on the 4D reconstruction process has been introduced and clinically evaluated \cite{Huizinga_2020}. 
However, this flow constraint is solely based on the geometry of the vessel tree and artifacts cannot be completely resolved.

To further constrain the reconstruction problem, physical models of blood flow and CA transport can be combined with the DSA measurements, such that reconstructions follow underlying physical laws and their associated hemodynamics. 
In particular, image-based blood flow simulations based on computational fluid dynamics (CFD) can be coupled with the imaging process to calculate the CA transport. Data consistency can be ensured by comparing  simulation-based virtual projection images and their respective real counterpart \cite{Boegel_2016, Endres_2012, Castro_2006, Durant_2008, Sun_2012}. 
One major advantage of the coupling of imaging and CFD simulations is the possibility to quantitatively evaluate 4D reconstruction algorithms, as real-world 4D contrast intensities cannot be measured with existing imaging modalities.
To lower computational demand and avoid a full 3D CFD simulation, reduced-order models, employing assumptions about the flow profile, have been proposed \cite{Waechter_2008}.
However, for all CFD coupling methods, vessel segments with high overlap or foreshortening artifacts must be excluded from the data consistency loss during the optimization. 
In addition, computationally expensive optimization must be conducted for each individual acquisition.

Summarizing, conventional 4D-DSA reconstruction methods still suffer from vessel overlap and vessel foreshortening artifacts. 
Reconstruction performance could be improved by constraining the solution to obey laws of fluid dynamics, however coupling measurements with numerical simulations is computationally expensive and a complex setup. 
In recent years, neural networks have been employed to solve inverse problems in computed tomography (CT), such as limited angle CT reconstruction \cite{wuerfl_2018} or denoising \cite{Wagner_2022}. These methods have become state-of-the-art in many applications. 
However, to the best of our knowledge, there exists no prior work on utilizing machine learning techniques for angiographic flow reconstruction in the vasculature.

In this work, we present a learning-based method for efficient and accurate time-resolved angiographic flow reconstruction.
Our contribution is a comprehensive pipeline for a learning-based reconstruction method that approximates the spatially averaged but time-resolved 1D+T CA concentration at all centerline points in the vasculature. 
These reconstructed 1D+T concentrations can be utilized for various applications, including serving as a surrogate for a full 4D reconstruction, correcting artifacts  (e.g., solving the vessel assignment problem in 4D-DSA tomographic backprojection) or directly running 1D hemodynamics estimation algorithms on them.
As vessel overlap and foreshortening are the cause of the ill-posedness, we model the reconstruction problem as an artifact correction problem, with the objective of correcting backprojected X-ray intensities.
To achieve this, a dataset of virtual rotational angiography acquisitions, based on 4D blood flow simulations and coupled CA transport simulations, is generated. A neural network is trained to efficiently approximate the artifact-free reconstruction by learning the mapping between backprojected intensities and ground truth CA concentrations.

\section{Method}
\subsection{Problem Description and Method Overview}
\label{sec:method:overview}

\begin{figure*}[!htb]
    \centering
    \scalebox{0.58}{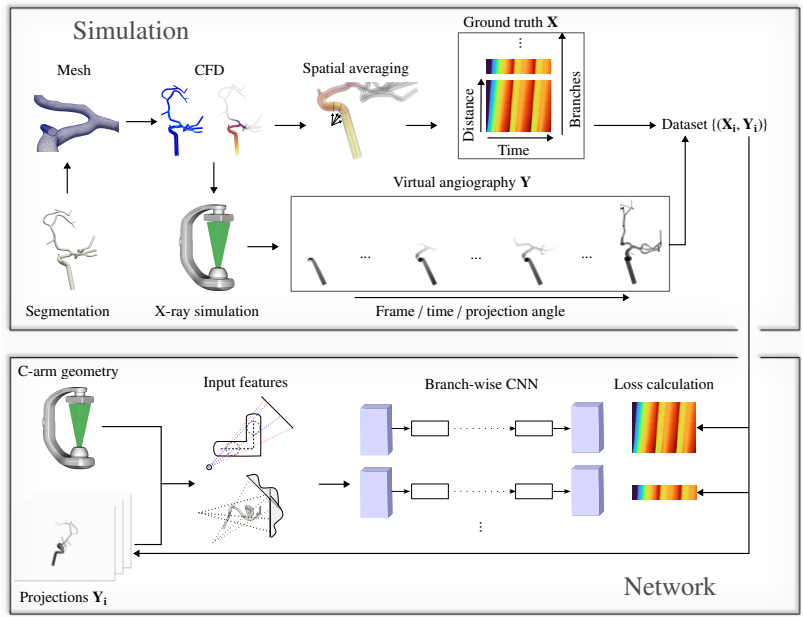}
    \caption{Overview of the learning-based method consisting of a simulation and a network part. \textbf{Simulation} Cerebral vessel tree surfaces extracted from segmentations are converted to volumetric polyhedral meshes. A computational fluid dynamics solver is employed to simulate hemodynamics and CA  transport for a set of boundary conditions. The simulated CA concentrations are spatially integrated for $P \in \mathbb{N}$ centerline slices and $T \in \mathbb{N}$ timesteps, resulting in $\mathbf{X} \in \mathbb{R}^{P \times T}$.
    Moreover, the X-ray C-arm acquisition process is simulated by computing the 2D projection image for each timepoint $\mathbf{Y} = (\mathbf{Y}_1, \dots, \mathbf{Y}_T | \, \mathbf{Y}_t \in \mathbb{R}^{H \times W})$. 
    A dataset $\{(\mathbf{X_i}, \mathbf{Y_i})\}$ of 1D+T concentrations and corresponding projection images is generated by simulating different flow and X-ray acquisition scenarios. 
    \textbf{Network} The projections $\mathbf{Y_i}$ and the C-arm geometry are utilized to compute backprojection, vessel overlap, and foreshortening input features for each centerline and time point. The centerline is split into branches that are processed individually by a convolutional neural network (CNN). The final loss is calculated with the ground truth CA concentration values $\mathbf{X_i}$.
    }
    \label{fig:method_overview}
\end{figure*}

The task of DSA flow reconstruction can be formulated as an inverse problem, where a time series of 2D projection images $\mathbf{Y} = (\mathbf{Y}_1, \dots, \mathbf{Y}_T \, | \, \mathbf{Y}_t \in \mathbb{R}^{H_y \times W_y})$ is utilized to reconstruct the original time series of 3D CA distributions $\mathbf{X}^\text{4D} = (\mathbf{X}^\text{3D}_1, \dots, \mathbf{X}^\text{3D}_T \, | \, \mathbf{X}^\text{3D}_t \in \mathbb{R}^{H_x \times W_x \times D_x})$. Each observed 2D filling state $\mathbf{Y}_t$ is linked to its 3D filling state $\mathbf{X}^\text{3D}_t$ by a forward process $A_t$ 
\begin{equation}
    \mathbf{Y}_t = A_t(\mathbf{X}^\text{3D}_t) + \epsilon_t \, ,
\end{equation}
in our case a cone-beam CT forward projection, and a noise term $\epsilon_t$. During the X-ray acquisition, each distinct 3D vessel tree filling state is observed as a 2D image on the detector. However, some attenuation information along the rays usually gets lost in the projection. 
When a ray passes through multiple vessels, the measured attenuation cannot be uniquely assigned to a single vessel. Instead, it may represent the sum of several contrasted vessels (vessel overlap).
Similarly, if the angle between ray direction and vessel direction is small, the measured attenuation at this point represents the total attenuation along the vessel instead of the attenuation along a single vessel cross section (vessel foreshortening). 

The filling states $\mathbf{X}^\text{3D}_t$ are linked by an underlying physical process $B$, that describes the CA flow between two timepoints $B(\mathbf{X}^\text{3D}_t) = \mathbf{X}^\text{3D}_{t+1}$. 
Prior knowledge about $B$ can be leveraged to constrain the space of possible states $\mathbf{X}^\text{3D}_t$ and alleviate the ill-posedness of a purely data-driven optimization.
Instead of explicitly integrating a model of $B$, e.g., by using a computationally expensive 3D CFD solver in the optimization procedure, we implicitly include knowledge about $B$ by training a neural network on a dataset constructed from CFD simulations.
Although simulating  a representative dataset is computationally expensive, a trained network can predict the reconstruction efficiently during inference.

To avoid the computational costs of predicting a time series of full voxel volumes, our method is designed to estimate the spatially integrated CA concentration $\mathbf{X} = (\mathbf{x}_1, \dots, \mathbf{x}_T \, | \, \mathbf{x}_t \in \mathbb{R}^P)$ at $P$ vessel centerline points and $T$ timesteps from a stack of projection images $\mathbf{Y} = (\mathbf{Y}_1, \dots, \mathbf{Y}_T \, | \, \mathbf{Y}_t \in \mathbb{R}^{H \times W})$. 
The inverse mapping $\mathbf{Y} \to \mathbf{X}$ is approximated using a neural network trained on a dataset consisting of several CA transport simulations and their corresponding virtual projection images. Once fully trained, the network can be applied to unseen DSA acquisitions and predict a solution efficiently. An overview of the method is presented in Fig. \ref{fig:method_overview}.

\subsection{Simulation and Data Generation}
\label{sec:data_acqui}

The dataset comprises simulated projection images and their corresponding 1D+T CA concentration maps, which serve as ground truth. 
The dataset is constructed using vessel trees that were segmented from rotational angiography data. 
In this study, we focus on cerebrovascular  geometries to evaluate the suitability of our method for neurovascular applications.
For this, we use a set of in-house segmentations as well as surface meshes  of the AneuX dataset \cite{aneux}.

\subsubsection{In-House Surface Mesh Pipeline}
The volumetric mesh generation consists of several stages. Initially, a multiscale vesselness filter is applied to the DSA reconstructions enhancing tubular structures \cite{Frangi_1998}. Next, a threshold-based algorithm generates a voxelized binary mask of the vascular structures. 
The segmented vessel tree is then pruned by removing noisy and small vessels to facilitate the simulation of hemodynamics.
Manual corrections are subsequently performed on the mask to ensure surfaces are free of artifacts, such as blending between two close vessels.
Finally, the binary mask is smoothed and converted into a triangular surface mesh.

\subsubsection{Aneurisk Surface Mesh Pipeline}
The AneuX dataset \cite{aneux} comprises surface meshes of cerebral vessel trees with aneurysms, which were previously segmented from 3D-DSA volumes.
Our study focuses on non-pathological cases and aneurysms are removed from the vascular trees.
For upstream aneurysms, the morphMan framework \cite{Kjeldsberg_2019, Kjeldsberg_2020} is utilized to remove aneurysms from the vasculature. 
For downstream aneurysms, the aneurysm and corresponding bifurcation is cut from the tree at the feeding artery.

\subsubsection{Volumetric Mesh Pipeline}
To model the inflow of blood and CA into the cerebral vasculature, the internal carotid artery (ICA) is cut at the cavernous segment and used as the flow inlet.
To ensure a developed flow and avoid backflow at the outlets (most distal vessel segments), flow extensions, with an approximate length of five times the respective vessel diameter, are added to the inlet and all outlets. As the extensions slightly increase the size of the tree and therefore add complexity to the problem (increased vascular overlap in projection directions), we treat the extensions as part of the tree.
Centerlines are calculated on the resulting mesh and a locally radius-adaptive tetrahedral mesh is generated \cite{Antiga_2008}. 
Five prismatic boundary layers are added to capture steep velocity gradients near the vessel walls. Subsequently, the polyhedral dual mesh is computed in \textit{OpenFOAM} (OpenFOAM, version 8, The OpenFOAM Foundation, London, United Kingdom) \cite{Weller_1998}.

\subsubsection{Physical Model}
The physical model describes the underlying physics and assumptions associated with the blood flow. Blood and CA are modeled as a Newtonian incompressible fluid with a kinematic viscosity $\nu$ of \SI{3.2e-6}{\meter\squared \per \second} and density of \SI{1060}{\kilogram\per\meter\cubed}.
The flow is assumed to be laminar. Mathematically, the model is described by the incompressible Navier-Stokes equations

\begin{equation}
\begin{split}
\frac{\partial \mathbf{u}}{\partial t}
    + \mathbf{u} \cdot \nabla \mathbf{u}
    &= - \nabla p + \nu \nabla^2 \mathbf{u}  \\
    \nabla \cdot \mathbf{u} &= 0 \, ,
\end{split} 
\end{equation}
where $\mathbf{u}$ denotes the fluid velocity and $p$ the pressure.
Vessel walls are modeled as rigid, with zero-gradient pressure and no-slip boundary conditions (BCs).

To augment the dataset, multiple flow scenarios are simulated for each geometry.
For this, inlet BCs are sampled from reported distributions in literature. 
The inflow waveform at the inlet is determined by three parameters: mean flow rate, cardiac cycle length, and age. 
Normalized inflow waveforms with varying cycle lengths are generated for young and elderly patients according to Ford et al. \cite{Ford_2005_ICA} and Hoi et al. \cite{Hoi_2010}. 
Cardiac cycle length and mean flow rate are sampled from a respective normal distribution with the reported mean and standard deviation \cite{Hoi_2010}, whereas the age is chosen uniformly as either young or elderly.  
Four different waveforms are generated for each geometry.

\subsubsection{Contrast Agent Model}
\label{sec:method:ca}
In a DSA acquisition, CA is injected to visualize vasculature downstream the injection point. The measured X-ray attenuation is therefore caused by a mixture of CA and blood with varying concentrations, which must be simulated.
To obtain a time series of 3D CA concentration distributions, a CA injection and transport model is coupled with the hemodynamics model.
Like in previous studies \cite{Ford_2005_virtual, Durant_2008, Sun_2012, Endres_2012, Waechter_2008}, it is assumed that the density difference between CA and blood is negligible and the transport is modeled using an advection-diffusion equation.
Hence, the transport equation of a passive tracer $c$ is given by
\begin{equation}
    \frac{\partial c} {\partial t}
    =  D \nabla^2 c - \mathbf{u} \cdot \nabla c \, ,
\end{equation}
where $D$ is a constant diffusion coefficient (independent of concentration) and $\mathbf{u}$ the underlying velocity field. The concentration $c$ is unit-less as it denotes the volume fraction of CA per unit volume.
In our case, we assume a maximum ICA injection flow rate $Q_\text{CA}^{\text{max}}(t)$ of \SI{2.5}{\milli \liter \per \second} which has been shown to result in an optimal quality of
4D-DSA temporal information \cite{Ruedinger_2019}. 
Due to resistances downstream, the total flow rate $Q_\text{T}(t)$ of the mixture can be modeled with a mixing factor $m$ \cite{Sun_2012, Mulder_2011}. The mixing factor determines the influence of the contrast injection on the total flow rate as described by the following equation
\begin{equation}
Q_\text{T}(t) = Q_\text{B}(t) + m \cdot Q_\text{CA}(t) \, ,
\end{equation}
where $Q_\text{B}(t)$ refers to the physiological blood flow rate before injection. As in previous studies \cite{maul_2023,Sun_2012}, we set the mixing factor to $0.3$. 
The compliance and resistance of the contrast flow through the catheter is modeled by an analogous electrical network consisting of a resistor and a capacitor, such that $Q_\text{CA}^{\text{max}}(t)$ is reached after some time lag.
Hence, the injection flow rate $Q_\text{CA}(t)$ is determined by 
\begin{equation}
    Q_\text{CA}(t) =
    \begin{cases}
    0  & t < T_\text{S} \\
    Q_\text{CA}^{\text{max}} \cdot (1 - \text{e}^{-(t-T_\text{S})/T_\text{L}}) & t \geq T_\text{S} 
    \end{cases} \, , 
\end{equation}
where $T_\text{S}$ denotes the injection start time and $T_\text{L}$ the lag \cite{Waechter_2008}. As common injection times are longer than three cardiac cycles, which is the simulated time span, the discharging can be neglected. 
In Fig. \ref{fig:flows} the waveforms are plotted for an example case.
Moreover, we assume that CA is injected into the ICA upstream and mixed uniformly with blood at the inlet. Hence, the concentration $c_\text{Inlet}(t)$ of CA at the inlet is defined by the ratio 
\begin{equation}
    c_\text{Inlet}(t) = \frac{Q_\text{CA}(t)}{Q_\text{T}(t)} \, .
\end{equation}

\subsubsection{Boundary Conditions and CFD Simulation}
\label{sec:sim_and_dataset}
\begin{figure}[!htb]
    \centering
    \includegraphics[width=0.9\linewidth]{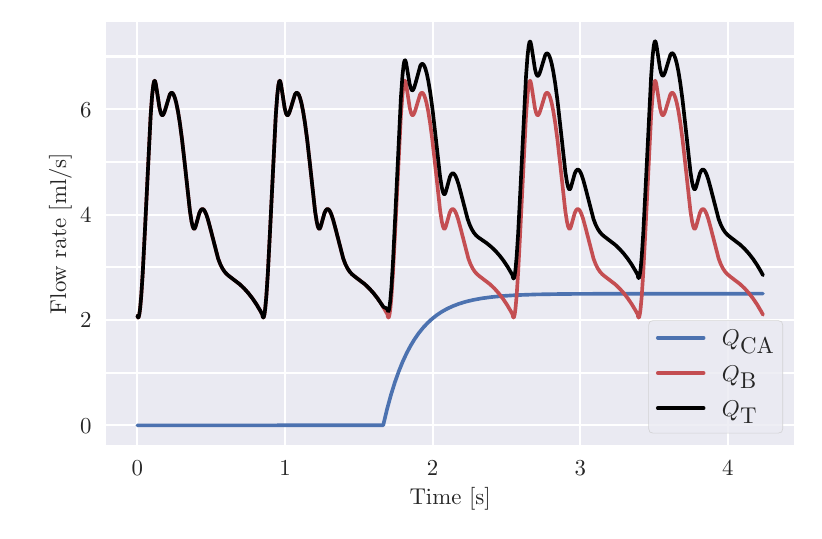}
    \caption{Plot visualizing the CA flow rate $Q_\text{CA}$, physiological blood flow rate $Q_{B}$ (assuming no injection), and the resulting total flow rate $Q_T$ with a mixing factor of $m=0.3$. The cardiac cycle waveform was generated for an elderly patient (secondary systole) with a mean $Q_B$ of \SI{4}{\milli\liter\per\second}.}
    \label{fig:flows}
    
\end{figure}

CFD simulations are conducted using the given mesh, physical model, and CA model. 
The outlet BCs are determined using a flow-spliting method, which prevents the usage of unrealistic zero pressure outlet BCs \cite{Chnafa_2018}. 
The algorithm starts at the inlet and calculates the flow split ratio between the branches at each bifurcation. 
In total, five cardiac cycles are simulated. The first cycle is used to wash out initial transient effects, whereas the second cycle reflects hemodynamics before contrast injection. 
The virtual injection of CA begins with the third cardiac cycle and continues until the end of the simulation, as depicted in Fig. \ref{fig:flows}. Second order schemes are selected for space and time discretization. An adaptive implicit time-stepping method with a maximum timestep of \SI{1}{\milli\second} is employed. 
All computations are conducted on a high performance computing cluster, utilizing 40 CPU cores per simulation. 

\subsubsection{Virtual Angiography Simulation}
To simulate X-ray imaging, virtual rotational angiographic projection images are computed.
In a clinical setting, the projection images of a mask run (without CA) and a fill run (CA injection) are subtracted to obtain images of the contrasted vessels only.
The remaining signal after subtraction is the X-ray attenuation of the blood and CA mixture inside the vasculature. Due to the significantly lower X-ray attenuation of blood compared to iodine-based CA, we perform a single material X-ray forward simulation.

At each timestep, the actual cone-beam CT acquisition is mimicked by forward projecting the CA density of the mixture using the corresponding projection geometry.
The forward projection is governed by the Lambert-Beer law, which describes X-ray attenuation behavior along rays. It is given by 
\begin{equation}
    \mathbf{Y}_t(\mathbf{u}) = \int I_0(E) \cdot \text{exp} \left(-\mu\left(E\right) 
    \int c\left(x,t\right)\text{d}l \right) 
    \text{d}E \, ,
\end{equation}
where $\mathbf{Y}_t(\mathbf{u}) $ is the intensity at detector pixel $\mathbf{u}$ and $I_0(E)$ the incident intensity for a given energy $E$.
The linear attenuation coefficient of the CA $\mu = \left(\mu / \rho\right)_\text{CA} \cdot \rho_\text{CA}$ can be decomposed into the mass attenuation coefficient $(\mu/ \rho)_\text{CA}$ and the CA density $\rho_\text{CA}$. 
We simulate the 2D-DSA images using DeepDRR \cite{Unberath_2018} and model $(\mu / \rho)_\text{CA}$ of the  \textit{Ultravist-300} (Bayer Vital GmbH, Leverkusen, Germany) CA with an iopromide concentration of \SI{623}{\milli\gram\per\milli\liter}. 
The CA mass attenuation coefficient $(\mu / \rho)_\text{CA}$ is calculated using the iopromide weight fraction $w_\text{IP}$ of the solution
\begin{equation}
\label{eq:concen_density}
    (\mu / \rho)_\text{CA} = w_\text{IP} (\mu/\rho)_\text{IP} + (1-w_\text{IP}) (\mu / \rho)_\text{W} \, ,
\end{equation}
where the remaining mass fraction $(1-w_\text{IP})$ is assumed to follow the attenuation behavior of water $(\mu / \rho)_\text{W}$.

Our simulations follow the common rotational 4D-DSA projection geometry. 
As the C-arm gantry rotates around the patient to record projections from multiple angles, the rotation is described by a primary angle $\alpha$ (measured around the cranio-caudal axis) and secondary tilt angle $\beta$. 
For each CFD and CA transport simulation, we generate nine projection series with varying starting primary angles $\alpha \in \{\SI{0}{\degree},\SI{55}{\degree},\SI{110}{\degree}\}$ and secondary angles $\beta \in \{\SI{-20}{\degree}, \SI{0}{\degree}, \SI{20}{\degree}\}$ of the C-arm to augment our dataset and simulate varying head poses.
The primary angle is increased by \SI{0.85}{\degree} for each timestep and 60 projection images per second are simulated.
\subsubsection{Spatial Averaging}

The spatially averaged concentration values for each centerline point are calculated, which are considered as the ground truth reconstruction.
Initially, centerlines are resampled at an inter-node distance of \SI{0.46}{\milli\meter}. Subsequently, at each centerline point, the mean contrast intensity is computed over the vessel cross-section perpendicular to the centerline for each timestep. Bifurcation regions are excluded from the computation as it is difficult to uniquely assign the concentration to the centerlines.

\subsection{Neural Network}
\begin{figure}[!htb]
    \centering
    \fontsize{6.3pt}{1pt}\selectfont
    \scalebox{0.95}{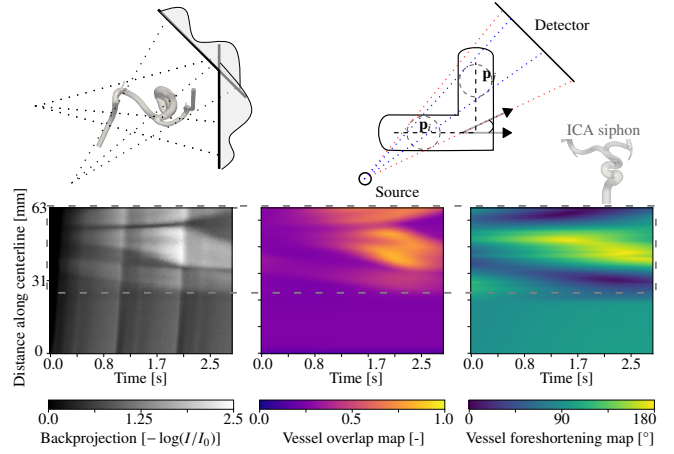}
    \caption{Visualization of the neural network input features for the ICA of a selected case. Measured intensities on the detector are backprojected to centerline points for each timestep. The vessel's overlap map between centerline points $\mathbf{p}_i$ and $\mathbf{p}_j$ for each timestep is determined by the overlap of the projected maximum inscribed spheres on the detector. The foreshortening map is the angle between the centerline normal vectors and the projection direction at each centerline point. Here, the siphone of the ICA introduces vessel overlap and foreshortening artifacts. }
    \label{fig:input_vis}
    
\end{figure}

We design a neural network model that predicts the average 1D+T concentrations at $P$ centerline points and $T$ timesteps given input features that are computed from the stack of projection images $\mathbf{Y}$.
We assume that a static 3D segmentation and centerlines of the vasculature are available, which can be derived from a common 3D-DSA reconstruction of $\mathbf{Y}$. 
We compute three input features, which is elaborated in detail below and exemplary results are visualized in Fig. \ref{fig:input_vis}. 

\subsubsection{Backprojection Input Feature}
A backprojection operation is employed to transform the measured intensities from projection space to reconstruction space. 
The operation utilizes information about the projection geometry from the image acquisition to forward project centerline points (calculated from the 3D segmentation of the vasculature) and subsequently sample bilinearly from the projection pixels at the respective locations.
 
To capture more volume of the vessel segment at a specific centerline point, virtual spheres are added around each centerline point in 3D reconstruction space and forward projected analogously. The mean of the sampled intensities is then calculated for each centerline point and its corresponding sphere points, resulting in an intensity matrix $\mathbf{I}\in \mathbb{R}^{P \times T}$ for all centerline points $P$ and timesteps $T$.

\subsubsection{Vessel Overlap Uncertainty Map Input Feature}

Waechter et al. \cite{Waechter_2008} define a binary uncertainty map, determining whether intensities are ignored in the case of  substantial vessel overlap or vessel foreshortening for the corresponding ray. 
Instead of using a binary mask, we compute an uncertainty value for each centerline point, which determines the degree of vessel overlap for a certain projection (and corresponding time step).
This results in a matrix $\mathbf{U} \in \mathbb{R}^{P \times T}$ that is later supplied to the network. 
Let $\tilde{a}_{i,t}$ denote the area of the forward projected maximum inscribed radius sphere \cite{Vmtk_2018} of the $i$-th centerline point at frame $t$. Also let $o_{i,j,t} \in \mathbb{R}_{\geq 0}$ denote the intersection area $\tilde{a}_{i,t} \cap \tilde{a}_{j,t}$. The uncertainty $u_{i,t} \in \mathbb{R}$ is then calculated as
\begin{equation}
    u_{i,t} = \frac{1}{\tilde{a}_{i,t} }\sum_{j=1}^{P} o_{i,j,t} \, ,
\end{equation}
where $u_{i,t} = P$ indicates that every centerline circle completely covers the circle of point $i$. On the other hand, $u_{i,t} = 1$ represents the scenario where there is no overlap between vascular structures.

\subsubsection{Vessel Foreshortening Uncertainty Map Input Feature}
The overlap uncertainty map takes the effect of foreshortening into account. This is because two centerline spheres that are close to each other may overlap in the projection for certain views. 
However, by using the known projection geometry and 3D vasculature, we can calculate a foreshortening uncertainty map $\mathbf{V} \in \mathbb{R}^{P \times T}$ to distinguish between vessel overlap and vessel foreshortening and provide additional information to the network.
Maximum foreshortening occurs when a vessel’s centerline is parallel to the intersecting ray, whereas the minimum occurs in the orthogonal case. 
Hence, for each projection $t$, we calculate the angle between the normalized ray $\mathbf{r}_{i,t} \in \mathbb{R}^3$ intersecting centerline point $i$ and the unit normal vector of the centerline point $\mathbf{n}_i \in \mathbb{R}^3$ using the inner product $v_{i,t} = \arccos{(\mathbf{r}_{i,t}^\intercal \mathbf{n}_{i})}$.

\subsubsection{Tree Decomposition}
The input tensor $\mathbf{Z} \in \mathbb{R}^{3 \times P \times T}$ is constructed for all centerline points in the tree by concatenating the backprojection, overlap uncertainty, and foreshortening uncertainty maps. 
We group centerline points by their respective branches (excluding bifurcation points) and denote the input of branch $b_i \in \{1,...,N\}$ as  $\mathbf{Z}^{b_i}$, where $N$ represents the number of branches. The 3D tensor $\mathbf{Z}$ (and analogously the output tensor $\mathbf{X}$) can then be decomposed into $\mathbf{Z} = ( \mathbf{Z}^{b_0}, \dots, \mathbf{Z}^{b_N}) $.
This decomposition allows to input each $\mathbf{Z}^{b_i}$ into a convolutional neural network (CNN) that is trained to learn the mapping $\mathbf{Z}^{b_i} \to \mathbf{X}^{b_i}$, where the spatial dimensions of the image can be interpreted as the distance along the centerline and time. 
After processing each branch with the network, the results are recombined in the tree data structure $\mathbf{X} = (\mathbf{X}^{b_0}, \dots, \mathbf{X}^{b_N})$. The branch decomposition can be regarded as a regularizing data augmentation technique, as a single tree contains multiple samples (similar to patch-wise processing), preventing overfitting on entire tree geometries by design.

\subsubsection{Network Architecture}
\begin{figure}[!htb]
    \centering
    \fontsize{11pt}{1pt}\selectfont
    \scalebox{0.6}{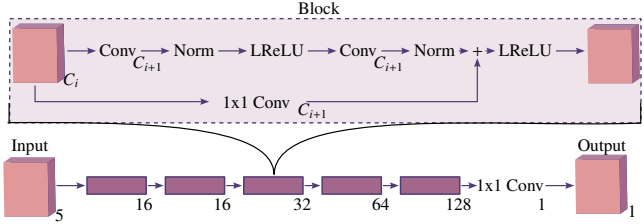}
    \caption{Visualization of the neural network architecture. It consists of five residual blocks with increasing number of channels $C_i$. A block applies two convolutional layers ($5\times5$ kernels), each followed by instance normalization, and a leaky rectified linear unit (LReLU). Before the output, a $1\times1$ convolution is applied. }
    \label{fig:architecture}
    
\end{figure}
We employ a lightweight CNN ResNet \cite{He_2016} architecture that consists of a repeating block of operations. Each block instance contains two convolutional layers, with each layer followed by instance normalization \cite{Ulyanov_2016} and a leaky rectified linear unit activation function. 
The entire architecture is depicted in Fig. \ref{fig:architecture}.

\subsubsection{Evaluation, Training, and Dataset Split}
 
The neural network is trained to regress the concentration maps $\mathbf{X}$ that capture the mean CA concentration $c(\mathbf{x}, t)$ across the vessel cross sections for each timestep. By definition, the concentration is unit-less as it describes the volume fraction of CA per unit volume, e.g., $0.5$ refers to a mixture where \SI{50}{\percent} of the volume is blood and \SI{50}{\percent} is CA.
However, using \eqref{eq:concen_density}, absolute quantitative information, such as iodine density or hounsfield units, can be determined. 
It should be noted that inferring absolute quantitative information from real X-ray images requires a calibration step, as recorded intensities on the detector are determined by the imaging procedure.

The reconstruction cases form a dataset that is split by geometry into training, validation, and test set. 
The training, validation, and test set consist of simulations from $21$, four, and five patient-based geometries, respectively. 
For each of the four CFD simulations of a geometry, nine different C-arm trajectories are simulated, leading to $4 \cdot 9 = 36$ samples per vessel tree.
In both training and validation set single CFD simulations that  did not finish within 24 hours and were discarded, resulting in $747$, $135$ and $180$ reconstruction cases in training, validation, and test set.
Additionally, the network is trained branch-wise, which increases the effective sample size. 
The network is trained to minimize the mean absolute error (MAE) 
\begin{equation}
\epsilon_\text{MAE} = \frac{1}{P \cdot T} \sum_{i=1}^P \sum_{j=1}^T |x_{ij} - \hat{x}_{ij}|
\end{equation}
between the ground truth concentration $\mathbf{X}$ and the predicted $\mathbf{\hat{X}}$ for $P$ centerline points and $T$ timesteps. 
The network is optimized for 300 epochs using the Adam optimizer \cite{Kingma_2015} and the weights resulting in the lowest validation loss are chosen for testing. 
During inference, we evaluate the absolute error and additionally the absolute percentage error 
\begin{equation}
\epsilon_\text{MAPE} = \frac{1}{P \cdot T} \sum_{i=1}^P \sum_{j=1}^T \frac{|x_{ij} - \hat{x}_{ij}|}{x_{ij}} \, .
\end{equation}
As the network parameters are optimized for the mean absolute error, large relative errors for small concentrations are not penalized in the loss, which can lead to extreme outliers. 
Hence, concentrations $ \leq 0.01$ are excluded from the relative error calculations.
It should be noted that we compute statistical measures (such as arithmetic mean, median, and standard deviation) of the errors and other variables on different data splits:
\begin{enumerate}
    \item Across branch-averaged and time-averaged errors to assess the performance spread on individual branches. 
    \item Across case-averaged and time-averaged errors to assess the performance spread across whole vessel trees.
\end{enumerate}
The mean errors are independent of the data split due to their linear computation. However, median and standard deviation may differ. 

\section{Results}
\subsection{Simulation Results}

\begin{figure*}[!htb]
    \centering
    \includegraphics[width=.65 \linewidth]{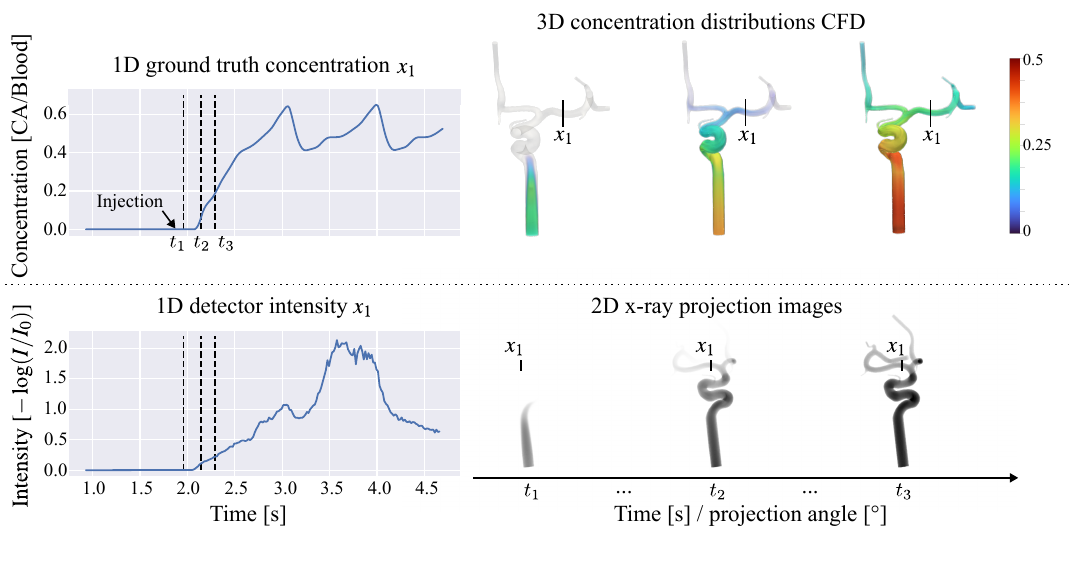}
    \caption{Result of our DSA simulation for an exemplary case. The ground truth concentration curve (CFD) and the detector intensity curve at position $x_1$ is visualized over time. Also, the 3D CFD vascular filling state at timesteps $t_1, t_2$, and $t_3$ in combination with the corresponding simulated projection images is depicted. Starting from approximately \SI{3.25}{\second}, the simulations include severe vessel foreshortening and overlap at $x_1$ for the respective projection angles.}
    \label{fig:cfd_result}
    
\end{figure*}

We exemplarily visualize the analysis of a simulated DSA acquisition (illustrated in Fig. \ref{fig:cfd_result}). 
Three timesteps of the 3D CFD vascular filling state and the corresponding projection images from the rotational scan are visualized. 
The spatially averaged CA concentration curve is plotted for a centerline point $x_1$ in the middle cerebral artery (MCA). 
Additionally, the corresponding intensities at the detector positions of the forward projected coordinates of $x_1$ are plotted. 
Although the attenuation of CA at $x_1$ shortly after the injection can be measured almost artifact-free, the vascular overlap and foreshortening corrupt the signal at later timesteps. 
Maximum information loss occurs when the direction of the MCA is parallel to the projection direction due to foreshortening and overlap with the anterior cerebral artery.

\subsection{Network Evaluation}

\subsubsection{Global Statistics}

\begin{figure}[!htb]
    \centering
        \includegraphics[width=0.85\linewidth]{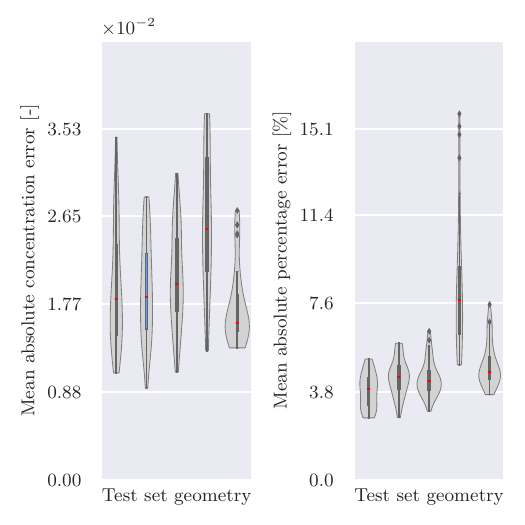}
        \caption{Violin plots visualizing the mean absolute error and mean absolute percentage distributions for the test set geometries (errors averaged for each case). The number of cases per geometry is $36$, as four different BCs and nine different C-arm trajectories are considered. }
        \label{fig:violin}
    \end{figure}

\begin{figure}[!htb]
        \centering
        \includegraphics[width=0.9\linewidth]{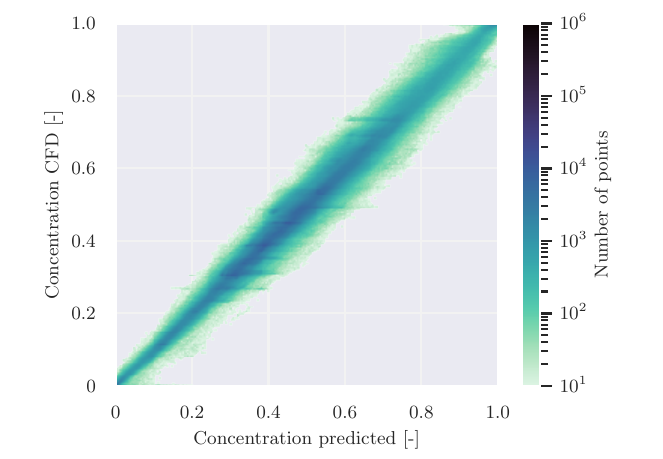}
        \caption{Regression plot of CA concentration values. For improved visualization, bins that contain less than $10$ elements are not plotted. Also, please note the logarithmic color map. The coefficient of determination is $0.98$ between predicted and ground truth CA concentration. }
        \label{fig:regression}
    
    \end{figure}

To assess the global performance of the trained model, we compute the case-averaged (over space and time) absolute and relative errors and calculate mean, median, and standard deviation across the $180$ cases. 
We receive $0.021, 0.019$, and $0.006$ for the absolute and \SI{5.18}{\percent}, \SI{4.52}{\percent}, and \SI{2.21}{\percent} for the relative error, respectively.
We measure a maximum mean absolute error of $0.037$ and a maximum mean absolute percentage error of \SI{15.8}{\percent}. 
To assess the performance for varying X-ray geometries and boundary conditions for a fixed vessel tree, we visualize the case-wise error distributions in a violin plot in Fig.  \ref{fig:violin}. 
The reconstruction of one particular vessel tree sample displays a larger mean absolute percentage error compared to all other samples. 
However, it should be noted that the network is not trained on the percentage error.
Further, we analyze the point-wise errors in a regression plot depicted in Fig.  \ref{fig:regression} (logarithmic colormap). No prediction bias was observed. However, smaller concentrations are predicted with less variance. The coefficient of determination ($R^2$) is $0.98$.


\subsubsection{Branch-Wise Analysis}
\begin{figure}[!htb]
    \centering
    \includegraphics[width=\linewidth]{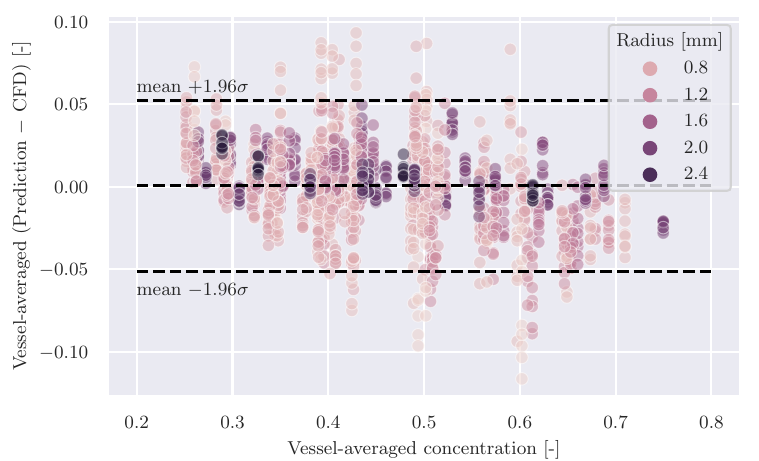}
    \caption{Bland-Altman plot visualizing the relationship between average vessel radius, vessel-averaged prediction error, and vessel-averaged concentration. Smaller vessels result in a higher error variance, whereas predictions for larger vessels result in a smaller error variance. The network tends to slightly overestimate low  concentrations and slightly underestimate high concentrations. }
    \label{fig:bland-altman}
    
\end{figure}

As the network is trained in a branch-wise manner and vessel morphology greatly influences the flow, the performance across all branches in the test set is analyzed.
In the Bland-Altman plot in Fig.  \ref{fig:bland-altman}, the vessel-averaged (averaged over space and time) error on the y-axis is plotted against the vessel-averaged concentration on the x-axis.
Further, the points are colored by vessel radius. We observe that vessels with bigger radii are predicted with less error than smaller ones and show little variance across the boundary conditions and X-ray geometry. The network tends to slightly overestimate smaller concentrations and slightly underestimate large concentrations.

\subsubsection{Influence of Physical and Geometric Parameters on the Model Prediction}

We analyze the influence of signal diffusivity, vessel overlap, geometric foreshortening, and branch flow rate on the prediction performance of the model.
Such analysis provides information about the robustness of the model across parameters and supports the identification of challenging reconstruction cases.
To avoid confounding factors, we analyze each branch individually to fix morphological parameters such as radii and length. 
For each branch, $36$ distinct samples exist in the dataset, as one tree is simulated with four different boundary conditions and nine different C-arm trajectories.
We compute the Pearson correlation coefficient of each branch-averaged physical parameter with the mean, median and standard deviation of the absolute percentage error for each branch individually (across $36$ cases) and calculate the mean correlation coefficients to get global statistical values for the entire test dataset.
The diffusivity is determined by calculating the negative standard deviation of the ground truth concentration maps. 

\begin{table}
\caption{Branch-wise Pearson correlation coefficients  of physical and geometric parameters with the mean, median, and standard deviation (Std) of the prediction error.}
\centering
\begin{tabular}{llll}
\toprule
Parameter    & Mean & Median & Std \\
\midrule
Diffusivity    & -0.12 & -0.05 & -0.21 \\
Overlap & 0.05 & 0.01 & 0.10\\
Foreshortening & 0.04 & 0.03 & 0.05\\
Flow rate & -0.09 & -0.03 & -0.19\\
\bottomrule
\end{tabular}
\label{tab:correlations}
\end{table}

We list the correlations in Table \ref{tab:correlations}. Overall, the numbers show little performance variations across the parameters, meaning that the network robustly predicts the concentrations for various settings.
We find that more diffusive concentration curves and cases with higher flow rates are predicted with a slightly smaller error on average.
The mean Pearson correlation coefficients for vessel overlap and vessel foreshortening are $0.05$ and $0.04$, respectively. Hence, only a small positive correlation between the degree of artifacts and prediction error exists, demonstrating robust network predictions.

\subsubsection{Qualitative Analysis}

\begin{figure*}[!htb]
    \centering
    \fontsize{6pt}{1pt}\selectfont
    \scalebox{0.8}{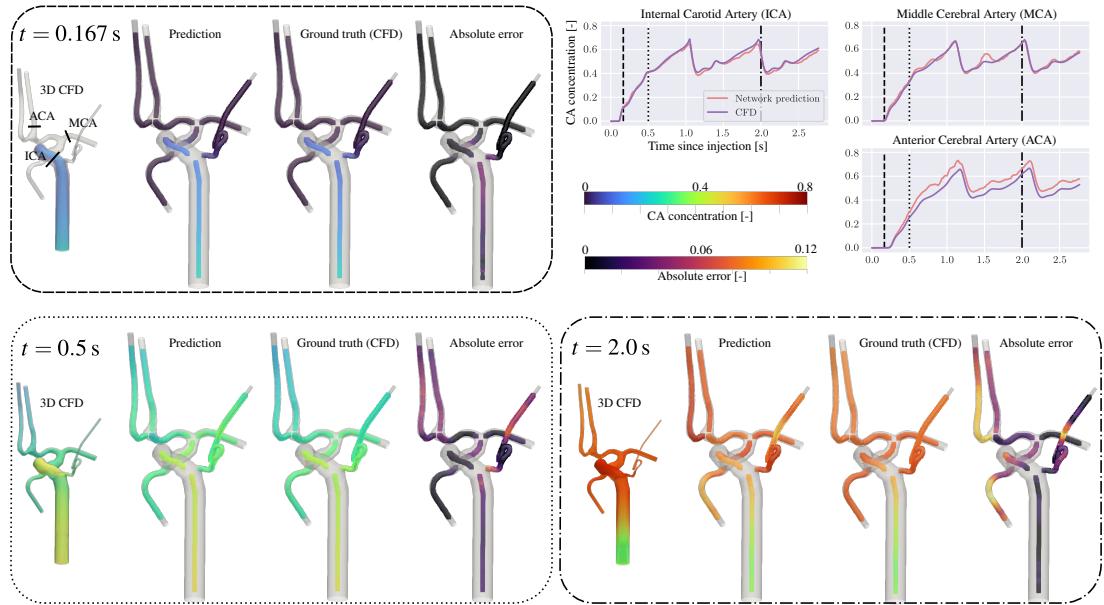}
    \caption{Qualitative analysis of a test set sample. Network concentration prediction, ground truth concentration, and absolute concentration error are depicted for three timesteps. Additionally, the concentration over time is plotted for a single cross section at the ICA, middle cerebral artery (MCA), and anterior cerebral artery (ACA), respectively. }
    \label{fig:qualitative_single_case}
\end{figure*}

We analyze an exemplary case from the test set in Fig. \ref{fig:qualitative_single_case} by comparing the network CA concentration prediction with the ground truth CFD CA concentration.
For this, three full tree visualizations are depicted that show a timestep with a low, medium and high CA concentration filling state, respectively.
We observe good agreement between predicted and ground truth concentrations for all three depicted timesteps. However, the error is higher for some distal vessel parts.
Moreover, the predicted and ground truth CA concentration over time is plotted for a cross section at the ICA, MCA, and anterior cerebral artery (ACA). 
Although prediction and CFD ground truth curves show good agreement for the ICA and MCA, the concentration in the ACA throughout time is slightly overestimated by the network. 

\section{Discussion}
\label{sec:discussion}
Time-resolved flow reconstruction from rotational angiography is an ill-posed inverse problem, where only a single projection image is available for each timestep. 
However, by incorporating prior knowledge about the fluid dynamics of blood and CA, overlap and foreshortening artifacts can be minimized, resulting in a  physically-plausible reconstruction. 
In this work, we present the first learning-based approach for time-resolved angiographic flow reconstruction, where physical prior knowledge is implicitly incorporated through training a neural network on a dataset of CFD simulations.
Our method utilizes a computationally efficient reduced-order 1D+T model, where the mean CA concentration for each cross-section along the centerline is considered.
Our evaluation shows that our trained model is able to reconstruct the CA concentration with a mean absolute error of \SI{0.02(0.02)}{} and a mean absolute percentage error of \SI{5.31(9.25)}{\percent} on the test set.

Previous approaches aimed at reducing overlap artifacts relied on classical approaches \cite{Huizinga_2020}, resulting in improved 4D image quality \cite{Xiang_2022}. However, remaining artifacts still required the assessment of 2D-DSA images for clinical decisions-making \cite{Keil_2022}. 
These works only qualitatively assessed the performance of their methods, making a direct comparison to our approach impossible.
In general, validating 4D angiographic reconstruction methods is difficult due to the inability to directly measure 4D CA distributions in the vasculature. Consequently, we can not compare our method against a gold standard.
Instead, we employed highly-resolved CFD simulations, which we believe provide a reliable representation of real CA distributions and enable quantitative evaluation of reconstruction performance.
Previous studies have demonstrated the capabilities of employing a CFD model \cite{Boegel_2016, Waechter_2008}. However, they are focused on hemodynamics estimation for simpler geometries and did not evaluate the CA reconstruction performance. 
Additionally, for these methods, optimization of model parameters requires extensive run times, simulations must be executed several times.
In contrast, our method requires many computational resources for dataset creation and neural network training, but is computationally efficient during inference.

Our method has some limitations. Due to the reduced-order 1D+T model, the neural network can only be applied for vessel structures and not for all pathological cases with complex hemodynamics inside more complicated vascular structures, such as aneurysms or arteriovenous malformations.
Similarly, we excluded bifurcation regions as it is challenging to map 3D bifurcation points to branching 1D centerlines.
Additionally, our method was so far not evaluated on real angiographic projection images. 
As a future step, we intend to conduct phantom studies, where flow probes can be inserted into a phantom and CFD simulations can be fitted to the measurements. 
Patient-specific CFD simulations, where the projection images are used to fit boundary conditions, are also an option. 
When applied to real projection images, a CA calibration is necessary, such that the recorded intensities can be mapped to a concentration value. However, these are complex setups and are beyond the scope of this research paper.

Our method could be extended for future research. 
The learning-based approach could be combined with aforementioned CFD parameter optimization methods. 
By using a 4D reconstruction, the data-consistency loss of the CA concentration can be directly computed in reconstruction space, avoiding projection space artifacts and computationally expensive forward projections \cite{Waechter_2008, Boegel_2016}. 
Moreover, physics-informed neural networks \cite{Raissi_2020, Shone_2023} could be employed to infer the hidden hemodynamics solely from the reconstructed concentrations, without requiring  ground truth CFD simulations. 
\section{Conclusion}
In conclusion, angiographic flow reconstruction has the potential to improve clinical decision-making for vascular abnormalities. We regard the flow reconstruction as a signal correction problem, where vessel overlap and vessel foreshortening artifacts are the cause of the ill-posedness.
In this study, we introduced the first neural network-based time-resolved angiographic flow reconstruction method, trained on CFD simulations, to correct these artifacts. 
Our approach showed promising performance, serving as a first step towards a research direction that leverages the combination of machine learning techniques and image-based blood flow simulations for angiographic flow reconstruction. 
\section*{Acknowledgment}
\sloppy
This project uses data from the AneuX morphology database, an open-access, multi-centric database combining data from three European projects: AneuX project (www.aneux.ch; @neurIST protocol v5; ethics autorisations Geneva BASEC PB\_2018‐00073; supported by the grant from the Swiss SystemsX.ch initiative, evaluated by the Swiss National Science Foundation), @neurIST project (www.aneurist.org; @neurIST protocol v1; ethics autorisations AmsterdamMEC 07-159, Barcelona2007-3507, Geneva CER 07-056, Oxfordshire REC AQ05/Q1604/162, Pècs RREC MC P 06 Jul 2007; supported by the 6th framework program of the European Commission FP6-IST-2004–027703) and Aneurisk (http://ecm2.mathcs.emory.edu/aneuriskweb/index).
The work of Fabian Wagner, Mareike Thies and Andreas Maier was supported by the European Research Council (ERC) under the European Union’s Horizon 2020 research and innovation program (ERC Grant No. 810316).
The work of Philipp Berg was supported by the German Research Foundation (SPP2311: project number 465189657) and by the German Federal Ministry of Education and Research within the Research Campus STIMULATE (grant no. 13GW0473A).

\section*{Disclaimer}
The concepts and information presented are based on research and are not commercially available.



\normalsize
\bibliography{refs}


\end{document}